\documentclass[11pt,a4paper]{article}
\usepackage[utf8]{inputenc}
\usepackage[T1]{fontenc}

\usepackage[hyperref]{acl2020}
\usepackage{times}

\usepackage{latexsym}
\usepackage{tabularx}
\usepackage{multicol}

\usepackage{microtype}

\aclfinalcopy %
\def\aclpaperid{97} %

\usepackage[normalem]{ulem}
\usepackage{enumitem}
\usepackage{array}
\usepackage{arydshln}
\usepackage{booktabs}
\usepackage{textcomp}
\usepackage[absolute]{textpos}

\usepackage{graphicx}

\def\ZKdel#1{\bgroup\markoverwith{\textcolor{green!60!black!100}{\rule[0.4ex]{2pt}{3pt}}}\ULon{#1}}

\def\ODdel#1{\bgroup\markoverwith{\textcolor{cyan!89!yellow!80!black!100}{\rule[0.4ex]{2pt}{3pt}}}\ULon{#1}}

\title{Evaluating Semantic Accuracy of Data-to-Text Generation\\ with Natural Language Inference}

\author{Ondřej Dušek \and Zdeněk Kasner\\
  Charles University, Faculty of Mathematics and Physics\\
  Institute of Formal and Applied Linguistics \\
  Prague, Czechia \\
  \texttt{\{odusek,kasner\}@ufal.mff.cuni.cz}
}

\date{}

\begin{document}
\maketitle
\begin{textblock*}{\textwidth}(2.5cm,1cm)
\noindent
In \emph{Proceedings of the 13th International Conference on Natural Language Generation (INLG)}, Online, December 2020.
\end{textblock*}

\begin{abstract}
A major challenge in evaluating data-to-text (D2T) generation is measuring the semantic accuracy of the generated text, i.e.\ %
checking if the output text contains all and only facts supported by the input data. %
We propose a new metric for evaluating the semantic accuracy of D2T generation %
based on a neural model pretrained for natural language inference (NLI). We use the NLI model to check textual entailment between the input data and the output text in both directions, allowing us to reveal omissions or hallucinations. 
Input data are converted to text for NLI using trivial templates. %
Our experiments on two recent D2T datasets show that our metric can achieve high accuracy in identifying erroneous system outputs. %
\end{abstract}

\section{Introduction}
\label{sec:intro}

Neural models may reduce the effort for building natural language generation (NLG) systems and produce very natural outputs, at the cost of limited control over the model outputs. State-of-the-art neural D2T models are prone to omitting or hallucinating facts \cite{gehrmann_end--end_2018,castro-ferreira-etal-2019-neural,duvsek2020evaluating}, which restricts their real-world deployment. Recognizing these errors is thus essential for %
proper system evaluation and further research in D2T generation.

In general, evaluating the semantic accuracy of D2T generation outputs requires full %
natural language understanding. Minor changes in wording 
may cause major differences in the meaning of the text, making it difficult for handcrafted heuristics to cover all edge cases. Human evaluation, on the other hand, is expensive and difficult to scale.

We note that the task of checking if a generated %
sentence includes/entails a particular fact is very close to the task of natural language inference (NLI). NLI is a sequence classification task which takes two inputs---a \textit{hypothesis} and a \textit{premise}---and produces one of the possible outputs: the hypothesis is \textit{entailed} by (follows from) the premise, \textit{contradicts} the premise, or their relation is \textit{neutral}. Recently, neural models for NLI \cite{zhang2019semantics,liu-etal-2019-multi,liu_roberta_2019} %
reached near-human levels of performance %
and NLI was used for evaluating the output of abstractive summarization systems \cite{maynez-etal-2020-faithfulness}. 

\begin{figure*}[t]
    \centering
    \includegraphics[width=0.9\textwidth]{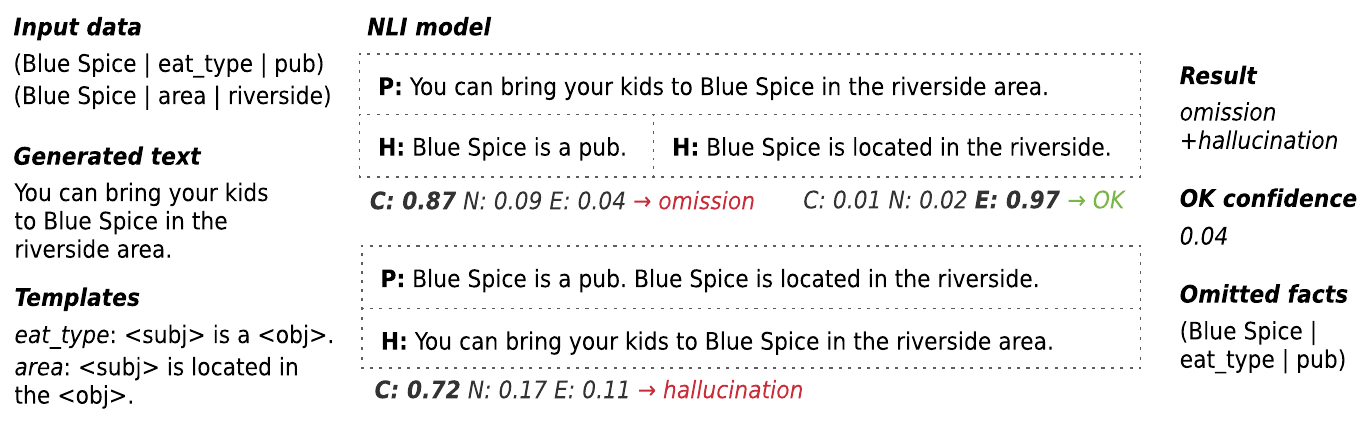}
    \caption{An example of evaluating the output from a D2T system with our metric. The generated text is used as a \textit{premise} (\textit{P}) to check for omissions and as a \textit{hypothesis} (\textit{H}) to check for hallucinations. The NLI model generates probabilities for \textit{contradiction} (\textit{C}), \textit{neutral} (\textit{N}) and \textit{entailment} (\textit{E}).}
    \label{fig:ex}
\end{figure*}

This brings a question: Can we use an NLI model for evaluating the semantic accuracy of D2T outputs?
The main idea of our method is to check with a general pretrained NLI model if the semantic information implied by the input data and the generated text is equal. We achieve this by using the NLI model to check for \textit{entailment} in two directions: By inferring input facts from the generated text we can check for \textit{omissions}, while the other direction allows us to check for \textit{hallucinations}.\footnote{This check in both directions is appropriate for D2T tasks that do not include content selection, which are the focus of our experiments in this paper. If the generator is supposed to select just some of the input facts to verbalize \cite[cf.~e.g.][]{wiseman_challenges_2017}, we can either only check for hallucinations or, if the content selection is explicit, perform a two-way check with the selected facts provided.} 
For instance, consider the two input facts from Figure~\ref{fig:ex}: \emph{(Blue Spice | eat\_type | pub)}, \emph{(Blue Spice | area | riverside)} and the generated text: “You can bring your kids to Blue Spice in the riverside area.” A NLI system should detect that the first fact is not entailed by the text (there is no mention of Blue Spice being a pub), but the text is also not entailed by the facts (the information about kids is hallucinated).%

Applying NLI for the D2T task brings a problem: %
The hypothesis for the standard NLI task is a natural language text, but the input for D2T generation is structured. However, we show that we can easily sidestep this issue by transforming the data into text using a trivial template for each fact.

We demonstrate %
that even without any human references or in-domain training and with minimal handcrafting, our approach achieves high accuracy (>90\%) on the E2E Challenge data \cite{duvsek2020evaluating}, competitive with scripts specifically handcrafted for the domain, and produces useful results (>75\% accuracy) %
on the more challenging %
WebNLG dataset \cite{gardent2017webnlg}. A manual error analysis shows that some instances marked as errors were in fact assessed correctly by our metric; we also identified a few major sources of errors that can be mitigated by in-domain tuning.
The experimental code for our metric is now available on GitHub.\footnote{\url{https://github.com/ufal/nlgi_eval}}

\section{Related Work}
\label{sec:related}
\paragraph{Automatic Evaluation of NLG} NLG outputs were traditionally evaluated by reference-based metrics measuring \emph{n}-gram overlap with a reference, such as BLEU \cite{papineni-etal-2002-bleu}, ROUGE \cite{lin-2004-rouge} and METEOR \cite{lavie_meteor:_2007}. Alternative, referenceless quality estimation metrics based on language model scores \cite{kann_sentence-level_2018} or linguistic features \cite{tian_treat_2018} focus on fluency and do not consider semantic accuracy. Recent works try to estimate NLG output quality with finetuned pretrained models \cite{zhou_learning_2020,zhang_bertscore:_2020,sellam_bleurt_2020}. The score from these models can capture some aspects of semantic accuracy, but only implicitly.

\paragraph{Semantic Accuracy}
To our knowledge, there is no generally accepted automatic metric for explicitly measuring semantic accuracy of NLG outputs. %
The closest commonly used metric is the \textit{slot error rate}, %
which is typically based on pattern matching tailored for a given dataset \cite{reed-etal-2018-neural,ijcai2019-437,duvsek2020evaluating}. Recently, \citet{goodrich_assessing_2019} introduced a metric based on training a neural model on named-entity recognition and fact extraction.

\paragraph{Faithful NLG}
Some recent neural NLG systems %
train specifically for semantic accuracy
\cite{nie-etal-2019-simple,tian2019sticking,kedzie-mckeown-2019-good}. Similarly to us, \citet{harkous2020have} use a pretrained neural model as a classifier to detect inaccurate output, finetuning the classifier on manually augmented domain-specific data.

Unlike previous works, we use a pretrained neural model finetuned for NLI which we do not further train on any domain-specific data.

\section{Method}
\label{sec:model}

\subsection{NLI Model}
\label{sec:nli-model}
We use pretrained RoBERTa \cite{liu_roberta_2019} as implemented in the Transformers library \cite{wolf_huggingfaces_2020} for our NLI model. Specifically, we use the \texttt{roberta-large-mnli}\footnote{\scalebox{0.95}[1.0]{\url{https://huggingface.co/roberta-large-mnli}}} checkpoint, which was finetuned on the MultiNLI dataset \cite{williams-etal-2018-broad}. We use the model \textit{as is}, without any further training. %
Given a premise text and a hypothesis text, the NLI model produces a probability distribution over three results: \textit{contradiction}, %
\textit{neutral} %
and \textit{entailment} (cf.~Section~\ref{sec:intro}). %
We consider a NLI check as passed if the probability for \textit{entailment} is the highest of the three.

\subsection{Data Preparation}
\label{sec:templates}

The input to our metric is a set of facts (the input for a D2T system) and the corresponding verbalization of these facts (the output from a D2T system). In our setup, the facts are RDF-like triples in the \textit{subject-predicate-object} form.%

We convert each triple 
to natural language using a trivial template. We consider two cases: 
\begin{enumerate}[nosep,label={(\arabic*)},leftmargin=14pt,labelwidth=12pt,labelsep=2pt]
    \item \emph{Default:} The templates can be handcrafted or extracted from the NLG systems' training data for each predicate.
    \item \emph{Backoff:} We use only a single, universal ``backoff'' template for all the facts, in the form: \emph{The \textless{}predicate\textgreater{} of \textless{}subject\textgreater{} is \textless{}object\textgreater{}}. 
\end{enumerate}
Hereinafter, a \textit{fact} refers to a template filled with the values from the triple.

\subsection{Evaluation Process}
\label{sec:eval-process}

The generated text is said to be correct if it mentions \textit{all} and \textit{only} the input facts.
We check if the text contains any omissions or hallucinations in two steps (see Figure~\ref{fig:ex} for an example): 
\begin{enumerate}[nosep,label={(\arabic*)},leftmargin=14pt,labelwidth=12pt,labelsep=2pt]
\item To check for omissions, we use the whole generated text as a premise and sequentially feed each fact as a hypothesis to the NLI model. Any failed NLI check is considered an omission. While we could use all concatenated facts in a single NLI check, our approach gives us further information about which facts are omitted. 
\item To check for hallucinations, we use a concatenation of all facts as a premise and feed the generated text as a hypothesis to the NLI model. If this NLI check fails, the text is considered to contain hallucination. This step cannot be split into simpler NLI checks.
\end{enumerate}
The final output of our metric is either 4-way (denoted as \textsc{Fine}): \emph{OK} (i.e., all NLI checks passed), \emph{omission}, \emph{hallucination} or \emph{omission+hallucination} (based on the failed checks), or 2-way (denoted as \textsc{Rough}) where the latter three results are collapsed into \emph{not\_OK}. The \textsc{Fine} 4-way output is more useful for system evaluation (we can distinguish whether the system tends to hallucinate or omit information). The \textsc{Rough} 2-way output corresponds more to a usage inside an NLG system for output reranking or filtering: any output that is \emph{not\_OK} should be penalized/\hspace{0mm}filtered out.
Additionally, we compute a \textit{confidence score} of the model as the minimum of all the entailment probabilities. %

\section{Experimental Setup}
\label{sec:experiments}

We experiment with two recent English data-to-text datasets with a triple-like format: WebNLG \cite{gardent2017webnlg} and E2E \cite{novikova-etal-2017-e2e}.\footnote{E2E data use attribute-value pairs relating to a restaurant; we convert them to triples where the restaurant is the subject.} Since both of them were used in shared tasks, sets of system outputs and measures of semantic accuracy are available (see Supplementary for details).

For WebNLG, we compare our metric with crowdsourced human ratings of semantic adequacy \cite{shimorina2019webnlg}. %
Human annotators used
a three-point Likert scale (1 = Incorrect, 2 = Medium, 3 = Correct) and answers are averaged over multiple annotators. In our experiments discussed in Section~\ref{sec:webnlg_results}, we consider a sentence correct if it achieved human rating 2.5 or higher (we also tried a threshold of 2.0, with slightly worse results).

For the E2E dataset, the challenge results were checked for semantic accuracy using a handcrafted automatic script \cite{duvsek2020evaluating},\footnote{While the E2E challenge did include crowdsourced evaluation of semantic accuracy, the results were unreliable, overestimating the number of errors \cite{duvsek2020evaluating}. Note that unlike our metric, such a handcrafted approach to evaluating semantic accuracy is only viable for limited domains such as E2E.%
} we therefore use this automatic script as the ground truth for evaluating our metric in Section~\ref{sec:e2e-results}.
We further use small sets of system outputs and human-written texts with expert annotation \citep[provided by][]{dusek_semantic_2019} to evaluate our approach against gold-standard annotation and to compare to existing semantic accuracy classifiers for E2E data in Section~\ref{sec:e2e-classifiers}.

We evaluate the \emph{Default} and \emph{Backoff} approaches to acquiring templates as described in Section~\ref{sec:templates}. The \emph{Default} setup works with one custom template per predicate type. For WebNLG, we obtained templates by delexicalizing human references for single-triple examples from WebNLG training data.\footnote{For each predicate, we choose randomly if more templates are found and use the backoff if no templates are found.} For E2E, we handcrafted 8 templates. 
The templates are filled with values from individual input triples and concatenated for multi-triple inputs as described in Section~\ref{sec:eval-process}. %

\section{Results Analysis}
\label{sec:results}

\begin{table}[t]
    \centering \small
    \begin{tabular}{l ccccc} \toprule
     & \textbf{A} & \textbf{R} & \textbf{P} & \textbf{F1} & $\mathbf{\rho}$ \\\midrule
    Default & 0.775 & 0.772 & 0.796 & 0.784 & 0.628 \\
    Backoff & 0.768 & 0.760 & 0.793 & 0.776 & 0.637 \\ \bottomrule
    \end{tabular}
    \caption{WebNLG dataset results, compared to crowdsourced human ratings (A = accuracy, R = recall, P = precision, F1 = F-measure, $\rho$ = Spearman correlation of confidence scores with human scores).}
    \label{tab:webnlg}
\end{table}

\begin{table}[t]
    \centering \small
    \begin{tabular}{l cccccc} \toprule
     & \textbf{Af} & \textbf{Ar} & \textbf{R} & \textbf{P} & \textbf{F1} \\\midrule
    Default & 0.911 & 0.933 & 0.895 & 0.910 & 0.903 \\
    Backoff & 0.846 & 0.874 & 0.913 & 0.768 & 0.834 \\ 
    \bottomrule
    \end{tabular}
    \caption{E2E dataset results, compared to the automatic evaluation script %
    (Af = \textsc{Fine} accuracy, Ar = \textsc{Rough} accuracy, R = recall, P = precision, F1 = F-measure).}
    \label{tab:e2e}
\end{table}

\begin{table*}[t]
    \centering\small
    \begin{tabular}{l c c ccc >{\hspace{5mm}} c c ccc}\toprule
    & \multicolumn{5}{c}{\bfseries Human-written (E2E training set)} & \multicolumn{5}{c}{\bfseries System outputs (TGen)} \\
     & \bf Af & \bf Ar & \bf R & \bf P & \bf F1 & \bf Af & \bf Ar & \bf R & \bf P & \bf F1 \\\midrule
    Slug2Slug aligner             & 0.685 & 0.765 & 0.550 & 0.800 & 0.652 & 0.995 & 1.000 & 1.000 & 1.000 & 1.000 \\
    E2E slot error script         & 0.820 & 0.885 & 1.000 & 0.777 & 0.874 & 0.995 & 0.995 & 1.000 & 0.950 & 0.974 \\
    TGen reranker     & 0.110 & 0.435 & 0.975 & 0.413 & 0.579 & 0.220 & 0.278 & 1.000 & 0.116 & 0.208 \\
    \midrule
    Default                       & 0.600 & 0.700 & 0.625 & 0.625 & 0.625 & 0.978 & 0.978 & 0.947 & 0.837 & 0.888 \\ %
    Backoff                       & 0.530 & 0.640 & 0.675 & 0.540 & 0.600 & 0.833 & 0.833 & 0.974 & 0.359 & 0.525 \\\bottomrule %
    \end{tabular}
    \caption{Semantic classifiers evaluated on expert human annotation on E2E data (see Table~\ref{tab:e2e} for metrics legend).} %
    \label{tab:e2e-slot-comparison}
\end{table*}

We evaluate our metric in terms of accuracy, precision, recall, and F1-measure (where \emph{not\_OK} samples are treated as positive since we focus on detecting errors). 
We additionally perform a manual error analysis on a random sample of 100 error examples for each dataset, i.e.\ examples where our metric gave a different assessment from the ground truth (provided by crowdsourced annotation for WebNLG and by a handcrafted classification script for E2E as described in Section~\ref{sec:experiments}). %
In general, the results are high above the random baseline (0.5 for the \textsc{Rough} metric and 0.25 for the \textsc{Fine} metric) but differ between the datasets, which we discuss below. %

\subsection{WebNLG Analysis}
\label{sec:webnlg_results}
The overall scores for the WebNLG dataset are summarized in Table \ref{tab:webnlg}. %
To further check whether the size of the input affects performance, we computed Spearman correlation of the number of input triples with metric errors. The resulting very low value of -0.05 ($p=$\,0.02, \emph{Default} setting) shows that the metric holds its performance even for more complex WebNLG examples. %
On the other hand, the overall scores show that our metric deviates quite a lot from the human judgments. Our manual error analysis indicates several reasons for that (see Supplementary for examples): (1) The annotation is somewhat noisy and using a threshold is not ideal---many correctly rendered outputs do not reach the 2.5 threshold (while some incorrect ones do). (2) Imprecise templates can confuse the NLI
(e.g., for the predicate \emph{nationality}, our extracted template is \emph{\textless{}subj\textgreater{} was \textless{}obj\textgreater{}}, which works well with values such as \emph{French}, but not with \emph{United States}). This is currently a weak point of our metric, as illustrated by the very small performance difference between the \emph{Default} and \emph{Backoff} setups; however, the issue can be mitigated by a better selection of the templates from training data, e.g.\ using language-model scoring. 
(3) The human annotators tend to give lower scores to accurate but ungrammatical or poorly organized texts. Our metric tends to rate these texts as \emph{OK}. 
Overall, our re-examination shows that almost half of the error examples (42 out of 100) were in fact correctly classified by our metric (i.e.\ their crowdsourced human annotation was incorrect), %
so the true performance is most likely higher than the reported numbers. %

The Spearman correlation of our model's confidence scores with the average human scores is around 0.63 ($p<$1e-10). This is similar to n-gram-based metrics on this data (\citealp{shimorina_human_2018} reports 0.59 for BLEU and 0.73 for METEOR), but unlike these metrics, our approach does not require human-written reference texts.

\subsection{E2E Analysis}
\label{sec:e2e-results}

The results for the E2E dataset (shown in Table \ref{tab:e2e}) are very good compared to the WebNLG dataset, with over 90\% agreement with the handcrafted script. This can be attributed to lower lexical variability and less noisy texts, as well as to the better quality of the handcrafted templates (the difference between the \emph{Default} and \emph{Backoff} setups is much more pronounced here).
Again, we observe only a very slight drop in performance for more complex E2E inputs (Spearman correlation of metric errors with the number of input triples is -0.08, $p<$1e-10 for the \emph{Default} setting).
The main issues identified by our error analysis are: 
(1) Problems in the interpretation of some values, e.g., \textit{price range=less than \textsterling{}20} is verbalized as ``cheap'' or \textit{family-friendly=no} as ``adult-only''. These cases are classified as \emph{not\_OK} by the NLI model.
(2) Missing or over-greedy patterns in the slot error script, causing annotation errors.
(3) Edge cases: some expressions cannot be interpreted in a straightforward way, e.g.\ ``high restaurant'' for \emph{pricerange=high} is deemed OK by the NLI but not by the slot error script.
(4) Expressions in the outputs that do not correspond to input facts, such as ``with full service'', are considered hallucinations by the NLI, but ignored by the slot error script.
Again, we consider about half of the error examples (45 out of 100) as correctly classified by our metric (see Supplementary for details), and thus our metric's performance is probably higher than the reported values due to erroneous annotation from the handcrafted script. 

\subsection{E2E MR Classifier Comparison}
\label{sec:e2e-classifiers}
We used expert-annotated E2E data samples (%
cf.~Section~\ref{sec:experiments}) to compare our approach to other accuracy classifiers in the E2E domain:
\begin{itemize}[nosep,leftmargin=10pt]
    \item \textbf{Slug2Slug slot aligner} \citep{juraska_deep_2018} is based on keyword matches. It is carefully tuned but not designed to detect hallucination; it only checks for presence of facts from the input MR.
    \item \textbf{E2E slot error script} (used in Section~\ref{sec:e2e-results}) is based on regular expressions; it is also able to detect irrelevant facts.
    \item \textbf{TGen reranker} is an LSTM-based model trained on the E2E training data to rerank outputs of the TGen system \cite{dusek_sequence--sequence_2016} based on their semantic accuracy. 
\end{itemize}

The results for all classifiers (in Table~\ref{tab:e2e-slot-comparison}) are much weaker on human-written data, which exhibit much more variability than system outputs.
The TGen reranker is very weak when required to detect all facts properly.
Our approach is slightly less precise than both handcrafted scripts, but the difference is small on system outputs (97.8\% vs. 99.5\% accuracy). If we disregard the value \emph{eatType=restaurant}, which is generally noisy, we get 76.5\% accuracy and 97.6\% recall on the human-written data. Moreover, our approach requires much less handcrafting and is more general.

\section{Conclusions and Future Work}
\label{sec:conclusion}
We described an automatic metric for evaluating semantic accuracy of D2T generation. With just a basic setup, without human references or training and with minimum handcrafting, our metric is able to detect omissions or hallucinations in generated texts, with results competitive with crowdsourced human ratings or handcrafted scripts customized for particular domains.

While our metric seems to scale well to more complex inputs in our experiments on the WebNLG and E2E data, we note that these datasets are still relatively limited. Further experiments are needed to evaluate this approach on long text generation and tasks where content selection is required, which we reserve for future work.
We also plan to integrate our metric as a reranker into an NLG system and apply small-scale in-domain finetuning in order to further improve results.
Following our findings from the error analysis on WebNLG, which showed that human ratings of semantic correctness are influenced by grammaticality, we would like to investigate the possibilities for combining our metric with a fluency/grammaticality checker \cite{kann_sentence-level_2018,tian_treat_2018}, as well as ways to better separate these two criteria in human evaluation. %

\section*{Acknowledgments}

We thank the anonymous reviewers for their helpful comments. This work was supported by the Charles University GAUK grant No.~140320, the SVV project No.~260575, and the Charles University project PRIMUS/19/SCI/10.

\bibliography{refs}

\begin{thebibliography}{32}
\expandafter\ifx\csname natexlab\endcsname\relax\def\natexlab#1{#1}\fi

\bibitem[{Castro~Ferreira et~al.(2019)Castro~Ferreira, van~der Lee, van
  Miltenburg, and Krahmer}]{castro-ferreira-etal-2019-neural}
Thiago Castro~Ferreira, Chris van~der Lee, Emiel van Miltenburg, and Emiel
  Krahmer. 2019.
\newblock \href {https://doi.org/10.18653/v1/D19-1052} {Neural data-to-text
  generation: A comparison between pipeline and end-to-end architectures}.
\newblock In \emph{Proceedings of the 2019 Conference on Empirical Methods in
  Natural Language Processing and the 9th International Joint Conference on
  Natural Language Processing (EMNLP-IJCNLP)}, pages 552--562, Hong Kong.

\bibitem[{Du{\v{s}}ek et~al.(2020)Du{\v{s}}ek, Novikova, and
  Rieser}]{duvsek2020evaluating}
Ond{\v{r}}ej Du{\v{s}}ek, Jekaterina Novikova, and Verena Rieser. 2020.
\newblock \href {https://doi.org/10.1016/j.csl.2019.06.009} {Evaluating the
  state-of-the-art of end-to-end natural language generation: The {E2E} {NLG}
  challenge}.
\newblock \emph{Computer Speech \& Language}, 59:123--156.

\bibitem[{Dušek et~al.(2019)Dušek, Howcroft, and
  Rieser}]{dusek_semantic_2019}
Ondřej Dušek, David~M Howcroft, and Verena Rieser. 2019.
\newblock \href {https://www.aclweb.org/anthology/W19-8652/} {Semantic {Noise}
  {Matters} for {Neural} {Natural} {Language} {Generation}}.
\newblock In \emph{Proceedings of the 12th {International} {Conference} on
  {Natural} {Language} {Generation} ({INLG} 2019)}, pages 421--426, Tokyo,
  Japan.

\bibitem[{Dušek and Jurčíček(2016)}]{dusek_sequence--sequence_2016}
Ondřej Dušek and Filip Jurčíček. 2016.
\newblock \href {https://aclweb.org/anthology/P16-2008} {Sequence-to-{Sequence}
  {Generation} for {Spoken} {Dialogue} via {Deep} {Syntax} {Trees} and
  {Strings}}.
\newblock In \emph{Proceedings of the 54th {Annual} {Meeting} of the
  {Association} for {Computational} {Linguistics} ({Volume} 2: {Short}
  {Papers})}, pages 45--51, Berlin.

\bibitem[{Gardent et~al.(2017)Gardent, Shimorina, Narayan, and
  Perez-Beltrachini}]{gardent2017webnlg}
Claire Gardent, Anastasia Shimorina, Shashi Narayan, and Laura
  Perez-Beltrachini. 2017.
\newblock \href {http://www.aclweb.org/anthology/W17-3518} {The {WebNLG}
  challenge: Generating text from {RDF} data}.
\newblock In \emph{Proceedings of the 10th International Conference on Natural
  Language Generation}, pages 124--133.

\bibitem[{Gehrmann et~al.(2018)Gehrmann, Dai, Elder, and
  Rush}]{gehrmann_end--end_2018}
Sebastian Gehrmann, Falcon~Z. Dai, Henry Elder, and Alexander~M. Rush. 2018.
\newblock \href {https://www.aclweb.org/anthology/W18-6505} {End-to-{End}
  {Content} and {Plan} {Selection} for {Data}-to-{Text} {Generation}}.
\newblock In \emph{Proceedings of the 11th {International} {Conference} on
  {Natural} {Language} {Generation}}, Tilburg, The Netherlands.

\bibitem[{Goodrich et~al.(2019)Goodrich, Rao, Saleh, and
  Liu}]{goodrich_assessing_2019}
Ben Goodrich, Vinay Rao, Mohammad Saleh, and Peter~J. Liu. 2019.
\newblock \href {https://doi.org/10.1145/3292500.3330955} {Assessing {The}
  {Factual} {Accuracy} of {Generated} {Text}}.
\newblock In \emph{{KDD}}, Anchorage, AK, USA.

\bibitem[{Harkous et~al.(2020)Harkous, Groves, and Saffari}]{harkous2020have}
Hamza Harkous, Isabel Groves, and Amir Saffari. 2020.
\newblock \href {http://arxiv.org/abs/2004.06577} {Have your text and use it
  too! end-to-end neural data-to-text generation with semantic fidelity}.
\newblock \emph{arXiv preprint arXiv:2004.06577}.

\bibitem[{Juraska et~al.(2018)Juraska, Karagiannis, Bowden, and
  Walker}]{juraska_deep_2018}
Juraj Juraska, Panagiotis Karagiannis, Kevin~K. Bowden, and Marilyn~A. Walker.
  2018.
\newblock \href {https://www.aclweb.org/anthology/N18-1014/} {A {Deep}
  {Ensemble} {Model} with {Slot} {Alignment} for {Sequence}-to-{Sequence}
  {Natural} {Language} {Generation}}.
\newblock In \emph{Proceedings of the 2018 {Conference} of the {North}
  {American} {Chapter} of the {Association} for {Computational} {Linguistics}:
  {Human} {Language} {Technologies}, {Volume} 1 ({Long} {Papers})}, pages
  152--162, New Orleans, LA, USA.

\bibitem[{Kann et~al.(2018)Kann, Rothe, and
  Filippova}]{kann_sentence-level_2018}
Katharina Kann, Sascha Rothe, and Katja Filippova. 2018.
\newblock \href {http://aclweb.org/anthology/K18-1031} {Sentence-{Level}
  {Fluency} {Evaluation}: {References} {Help}, {But} {Can} {Be} {Spared}!}
\newblock In \emph{Proceedings of the 22nd {Conference} on {Computational}
  {Natural} {Language} {Learning}}, pages 313--323, Brussels, Belgium.

\bibitem[{Kedzie and McKeown(2019)}]{kedzie-mckeown-2019-good}
Chris Kedzie and Kathleen McKeown. 2019.
\newblock \href {https://doi.org/10.18653/v1/W19-8672} {A good sample is hard
  to find: Noise injection sampling and self-training for neural language
  generation models}.
\newblock In \emph{Proceedings of the 12th International Conference on Natural
  Language Generation}, pages 584--593, Tokyo, Japan.

\bibitem[{Lavie and Agarwal(2007)}]{lavie_meteor:_2007}
Alon Lavie and Abhaya Agarwal. 2007.
\newblock \href {https://www.aclweb.org/anthology/W07-0734} {Meteor: {An}
  {Automatic} {Metric} for {MT} {Evaluation} with {High} {Levels} of
  {Correlation} with {Human} {Judgments}}.
\newblock In \emph{Proceedings of the {Second} {Workshop} on {Statistical}
  {Machine} {Translation}}, pages 228--231, Prague, Czech Republic. Association
  for Computational Linguistics.

\bibitem[{Lin(2004)}]{lin-2004-rouge}
Chin-Yew Lin. 2004.
\newblock \href {https://www.aclweb.org/anthology/W04-1013} {{ROUGE}: A package
  for automatic evaluation of summaries}.
\newblock In \emph{Text Summarization Branches Out}, pages 74--81, Barcelona,
  Spain.

\bibitem[{Liu et~al.(2019{\natexlab{a}})Liu, He, Chen, and
  Gao}]{liu-etal-2019-multi}
Xiaodong Liu, Pengcheng He, Weizhu Chen, and Jianfeng Gao. 2019{\natexlab{a}}.
\newblock \href {https://doi.org/10.18653/v1/P19-1441} {Multi-task deep neural
  networks for natural language understanding}.
\newblock In \emph{Proceedings of the 57th Annual Meeting of the Association
  for Computational Linguistics}, pages 4487--4496, Florence, Italy.

\bibitem[{Liu et~al.(2019{\natexlab{b}})Liu, Ott, Goyal, Du, Joshi, Chen, Levy,
  Lewis, Zettlemoyer, and Stoyanov}]{liu_roberta_2019}
Yinhan Liu, Myle Ott, Naman Goyal, Jingfei Du, Mandar Joshi, Danqi Chen, Omer
  Levy, Mike Lewis, Luke Zettlemoyer, and Veselin Stoyanov. 2019{\natexlab{b}}.
\newblock \href {http://arxiv.org/abs/1907.11692} {{RoBERTa}: {A} {Robustly}
  {Optimized} {BERT} {Pretraining} {Approach}}.
\newblock \emph{arXiv preprint arXiv:1907.11692}.

\bibitem[{Maynez et~al.(2020)Maynez, Narayan, Bohnet, and
  McDonald}]{maynez-etal-2020-faithfulness}
Joshua Maynez, Shashi Narayan, Bernd Bohnet, and Ryan McDonald. 2020.
\newblock \href {https://doi.org/10.18653/v1/2020.acl-main.173} {On
  faithfulness and factuality in abstractive summarization}.
\newblock In \emph{Proceedings of the 58th Annual Meeting of the Association
  for Computational Linguistics}, pages 1906--1919, Online.

\bibitem[{Mi et~al.(2019)Mi, Huang, Zhang, and Faltings}]{ijcai2019-437}
Fei Mi, Minlie Huang, Jiyong Zhang, and Boi Faltings. 2019.
\newblock \href {https://doi.org/10.24963/ijcai.2019/437} {Meta-learning for
  low-resource natural language generation in task-oriented dialogue systems}.
\newblock In \emph{Proceedings of the Twenty-Eighth International Joint
  Conference on Artificial Intelligence, {IJCAI-19}}, pages 3151--3157.

\bibitem[{Nie et~al.(2019)Nie, Yao, Wang, Pan, and Lin}]{nie-etal-2019-simple}
Feng Nie, Jin-Ge Yao, Jinpeng Wang, Rong Pan, and Chin-Yew Lin. 2019.
\newblock \href {https://doi.org/10.18653/v1/P19-1256} {A simple recipe towards
  reducing hallucination in neural surface realisation}.
\newblock In \emph{Proceedings of the 57th Annual Meeting of the Association
  for Computational Linguistics}, pages 2673--2679, Florence, Italy.

\bibitem[{Novikova et~al.(2017)Novikova, Du{\v{s}}ek, and
  Rieser}]{novikova-etal-2017-e2e}
Jekaterina Novikova, Ond{\v{r}}ej Du{\v{s}}ek, and Verena Rieser. 2017.
\newblock \href {https://doi.org/10.18653/v1/W17-5525} {The {E}2{E} dataset:
  New challenges for end-to-end generation}.
\newblock In \emph{Proceedings of the 18th Annual {SIG}dial Meeting on
  Discourse and Dialogue}, pages 201--206, Saarbr{\"u}cken, Germany.

\bibitem[{Papineni et~al.(2002)Papineni, Roukos, Ward, and
  Zhu}]{papineni-etal-2002-bleu}
Kishore Papineni, Salim Roukos, Todd Ward, and Wei-Jing Zhu. 2002.
\newblock \href {https://doi.org/10.3115/1073083.1073135} {{B}leu: a method for
  automatic evaluation of machine translation}.
\newblock In \emph{Proceedings of the 40th Annual Meeting of the Association
  for Computational Linguistics}, pages 311--318, Philadelphia, Pennsylvania,
  USA.

\bibitem[{Reed et~al.(2018)Reed, Oraby, and Walker}]{reed-etal-2018-neural}
Lena Reed, Shereen Oraby, and Marilyn Walker. 2018.
\newblock \href {https://doi.org/10.18653/v1/W18-6535} {Can neural generators
  for dialogue learn sentence planning and discourse structuring?}
\newblock In \emph{Proceedings of the 11th International Conference on Natural
  Language Generation}, pages 284--295, Tilburg University, The Netherlands.

\bibitem[{Sellam et~al.(2020)Sellam, Das, and Parikh}]{sellam_bleurt_2020}
Thibault Sellam, Dipanjan Das, and Ankur~P. Parikh. 2020.
\newblock \href {https://www.aclweb.org/anthology/2020.acl-main.704} {{BLEURT}:
  {Learning} {Robust} {Metrics} for {Text} {Generation}}.
\newblock In \emph{Proceedings of the 58th {Annual} {Meeting} of the
  {Association} for {Computational} {Linguistics}}, pages 7881--7892, Online.

\bibitem[{Shimorina(2018)}]{shimorina_human_2018}
Anastasia Shimorina. 2018.
\newblock \href {http://arxiv.org/abs/1805.11474} {Human vs {Automatic}
  {Metrics}: on the {Importance} of {Correlation} {Design}}.
\newblock In \emph{{WiNLP} {Workshop}}, New Orleans, LA, USA.

\bibitem[{Shimorina et~al.(2019)Shimorina, Gardent, Narayan, and
  Perez-Beltrachini}]{shimorina2019webnlg}
Anastasia Shimorina, Claire Gardent, Shashi Narayan, and Laura
  Perez-Beltrachini. 2019.
\newblock \href
  {https://webnlg-challenge.loria.fr/files/human-eval-outline-v2.pdf} {{WebNLG}
  challenge: Human evaluation results}.
\newblock Technical report, LORIA.

\bibitem[{Tian et~al.(2019)Tian, Narayan, Sellam, and
  Parikh}]{tian2019sticking}
Ran Tian, Shashi Narayan, Thibault Sellam, and Ankur~P Parikh. 2019.
\newblock \href {https://arxiv.org/abs/1910.08684} {Sticking to the facts:
  Confident decoding for faithful data-to-text generation}.
\newblock \emph{arXiv preprint arXiv:1910.08684}.

\bibitem[{Tian et~al.(2018)Tian, Douratsos, and Groves}]{tian_treat_2018}
Ye~Tian, Ioannis Douratsos, and Isabel Groves. 2018.
\newblock \href {http://aclweb.org/anthology/W18-6512} {Treat the system like a
  human student: {Automatic} naturalness evaluation of generated text without
  reference texts}.
\newblock In \emph{Proceedings of the 11th {International} {Conference} on
  {Natural} {Language} {Generation}}, pages 109--118, Tilburg University, The
  Netherlands.

\bibitem[{Williams et~al.(2018)Williams, Nangia, and
  Bowman}]{williams-etal-2018-broad}
Adina Williams, Nikita Nangia, and Samuel Bowman. 2018.
\newblock \href {https://doi.org/10.18653/v1/N18-1101} {A broad-coverage
  challenge corpus for sentence understanding through inference}.
\newblock In \emph{Proceedings of the 2018 Conference of the North {A}merican
  Chapter of the Association for Computational Linguistics: Human Language
  Technologies, Volume 1 (Long Papers)}, pages 1112--1122, New Orleans,
  Louisiana.

\bibitem[{Wiseman et~al.(2017)Wiseman, Shieber, and
  Rush}]{wiseman_challenges_2017}
Sam Wiseman, Stuart~M. Shieber, and Alexander~M. Rush. 2017.
\newblock \href {https://aclweb.org/anthology/D17-1239} {Challenges in
  {Data}-to-{Document} {Generation}}.
\newblock In \emph{Proceedings of the 2017 {Conference} on {Empirical}
  {Methods} in {Natural} {Language} {Processing}}, pages 2243--2253,
  Copenhagen, Denmark.

\bibitem[{Wolf et~al.(2020)Wolf, Debut, Sanh, Chaumond, Delangue, Moi, Cistac,
  Rault, Louf, Funtowicz, Davison, Shleifer, von Platen, Ma, Jernite, Plu, Xu,
  Scao, Gugger, Drame, Lhoest, and Rush}]{wolf_huggingfaces_2020}
Thomas Wolf, Lysandre Debut, Victor Sanh, Julien Chaumond, Clement Delangue,
  Anthony Moi, Pierric Cistac, Tim Rault, Rémi Louf, Morgan Funtowicz, Joe
  Davison, Sam Shleifer, Patrick von Platen, Clara Ma, Yacine Jernite, Julien
  Plu, Canwen Xu, Teven~Le Scao, Sylvain Gugger, Mariama Drame, Quentin Lhoest,
  and Alexander~M. Rush. 2020.
\newblock \href {http://arxiv.org/abs/1910.03771} {{HuggingFace}'s
  {Transformers}: {State}-of-the-art {Natural} {Language} {Processing}}.
\newblock \emph{arXiv preprint arXiv:1910.03771}.

\bibitem[{Zhang et~al.(2020{\natexlab{a}})Zhang, Kishore, Wu, Weinberger, and
  Artzi}]{zhang_bertscore:_2020}
Tianyi Zhang, Varsha Kishore, Felix Wu, Kilian~Q. Weinberger, and Yoav Artzi.
  2020{\natexlab{a}}.
\newblock \href {http://arxiv.org/abs/1904.09675} {{BERTScore}: {Evaluating}
  {Text} {Generation} with {BERT}}.
\newblock In \emph{{ICLR}}, Online.

\bibitem[{Zhang et~al.(2020{\natexlab{b}})Zhang, Wu, Zhao, Li, Zhang, Zhou, and
  Zhou}]{zhang2019semantics}
Zhuosheng Zhang, Yuwei Wu, Hai Zhao, Zuchao Li, Shuailiang Zhang, Xi~Zhou, and
  Xiang Zhou. 2020{\natexlab{b}}.
\newblock \href {https://arxiv.org/abs/1909.02209} {Semantics-aware {BERT} for
  language understanding}.
\newblock In \emph{Thirty-Fourth AAAI Conference on Artificial Intelligence
  (AAAI-2020)}.

\bibitem[{Zhou and Xu(2020)}]{zhou_learning_2020}
Wangchunshu Zhou and Ke~Xu. 2020.
\newblock \href {http://arxiv.org/abs/2002.05058} {Learning to {Compare} for
  {Better} {Training} and {Evaluation} of {Open} {Domain} {Natural} {Language}
  {Generation} {Models}}.
\newblock In \emph{{AAAI}}, New York, NY, USA.

\end{thebibliography}
\bibliographystyle{acl_natbib}

\clearpage
\appendix

\onecolumn

\noindent{\large\bfseries Supplementary Material: Evaluating Semantic Accuracy of Data-to-Text Generation with Natural Language Inference}

\section*{Dataset details}

The WebNLG data used for our experiments is the subset of NLG system outputs used for human evaluation \cite{shimorina2019webnlg}\footnote{The data is available at \url{https://gitlab.com/webnlg/webnlg-human-evaluation}. We used the file \texttt{all\_data\_final\_averaged.csv}.} -- 223 sampled data inputs from the WebNLG 2017 test set with 10 different NLG system outputs for each input, i.e., 2,230 instances in total.

For the main E2E experiments (with the slot error script as ground truth in Section~\ref{sec:e2e-results}), we used the full set of primary system outputs on the whole E2E test set \cite{duvsek2020evaluating} -- 21 outputs for 630 input data items each, i.e., 13,230 instances in total.\footnote{The data are available from \url{https://github.com/tuetschek/e2e-eval}. We used all the files under \texttt{system\_outputs/primary}.}
For semantic classifier comparison in Section~\ref{sec:e2e-classifiers}, we used two expert-annotated sets provided to us by \citet{dusek_semantic_2019}, who used them for their slot error script evaluation: 200 instances from the E2E training set (the human-written texts were reannotated as many of them did not reflect the original input properly) and 400 outputs of different variants of the TGen NLG system \cite{dusek_sequence--sequence_2016}.

\section*{WebNLG Error Analysis}

We checked 100 randomly sampled examples from the WebNLG data where our approach and the crowdsourced human annotation gave different results. We identified 51 cases where the crowdsourced human annotation was indeed correct, 42 where the human annotation was incorrect but our NLI-based approach provided a correct result, and 7 cases where both annotations were incorrect or where it was not possible to unambiguously decide between the two.

The main error types identified, with counts and examples, are as follows (note that the analyzed examples may belong to multiple or none of the above classes):
\begin{enumerate}[label={\bf (\arabic*)},leftmargin=14pt,labelwidth=12pt,labelsep=2pt]
    \item \textbf{Annotation problems} -- 22 counts. Example:

\begin{table}[h]
    \centering \small
    \renewcommand{\arraystretch}{1.25}
    \begin{tabular}{p{0.45\textwidth} p{0.45\textwidth}}
    \textbf{Data} &  \textbf{Templates} \\ 
     \textit{1 Decembrie 1918 University | state | Alba}  
     & 1 Decembrie 1918 University stands in the state of Alba. \\
     
    \rule{0pt}{3ex}\textbf{Text} \\  \multicolumn{2}{p{0.9\textwidth}}{1 decembrie 1918 university is in the state of alba.}\\
    
     \rule{0pt}{3ex}\textbf{Human Output}  &  \textbf{NLI Output} \\  2.33 \textit{(=not\_OK)} & \textit{OK} \\  
    \rule{0pt}{3ex}\textbf{Commentary} & \\
    
    \multicolumn{2}{p{0.9\textwidth}} {The sentence is OK, but the human score is slightly below the threshold for no apparent reason.}
     \end{tabular}
     
    \label{tab:supp-webnlg-1}
\end{table}

    \item \textbf{Inaccurate templates} -- 22 counts. Example:

\begin{table}[h]
    \centering \small
    \renewcommand{\arraystretch}{1.25}
    \begin{tabular}{p{0.45\textwidth} p{0.45\textwidth}}
    \textbf{Data} &  \textbf{Templates} \\ 
     \textit{Aenir | language | English language}  
     & One of the languages of Aenir is English language. \\
     
    \rule{0pt}{3ex}\textbf{Text} \\  \multicolumn{2}{p{0.9\textwidth}}{aenir is written in english.}\\
    
     \rule{0pt}{3ex}\textbf{Human Output}  &  \textbf{NLI Output} \\  3 \textit{(=OK)} & \textit{hallucination} \\  
    \rule{0pt}{3ex}\textbf{Commentary} & \\
    
    \multicolumn{2}{p{0.9\textwidth}} {The sentence is OK, but the template is not specific enough for a literary work, which leads the NLI to assume this is a hallucination.}
     \end{tabular}
     
    \label{tab:supp-webnlg-2}
\end{table}

\newpage
    \item \textbf{Disfluent or disorganized sentences} -- 14 counts. Example:

\begin{table}[h]
    \centering \small
    \renewcommand{\arraystretch}{1.25}
    \begin{tabular}{p{0.45\textwidth} p{0.45\textwidth}}
    \textbf{Data} &  \textbf{Templates} \\ 
     \textit{Spain | leaderName | Felipe VI of Spain}  
     & the name of the leader of Spain is Felipe VI of Spain. \\
     \textit{Ajoblanco | region | Andalusia} & Ajoblanco can be found Andalusia. \\
     
     \textit{Andalusia | leaderName | Susana Díaz} & the name of the leader of  Andalusia is Susana Díaz. \\
     
     \textit{Ajoblanco | country | Spain} & Ajoblanco is situated within Spain. \\
     
     \textit{Spain | demonym | Spaniards} & The demonym for natives of Spain is Spaniards. \\
    \rule{0pt}{3ex}\textbf{Text} \\  \multicolumn{2}{p{0.9\textwidth}}{ajoblanco is a spanish dish that originates andalusia , where the  leader is susana diaz. felipe vi is the leader of spain , where the  people that live there are called spaniards.}\\
    
     \rule{0pt}{3ex}\textbf{Human Output}  &  \textbf{NLI Output} \\  2.33 \textit{(=not\_OK)} & \textit{OK} \\  
    \rule{0pt}{3ex}\textbf{Commentary} & \\
    
    \multicolumn{2}{p{0.9\textwidth}} {The sentence is OK, but not completely fluent and not very well organized (some of the facts are not very related). This probably leads to the lower human score, but doesn't cause any problems to the NLI.}
     \end{tabular}
     
    \label{tab:supp-webnlg-3}
\end{table}

    \item \textbf{Value format errors} -- 7 counts. Example:

\begin{table}[h]
    \centering \small
    \renewcommand{\arraystretch}{1.25}
    \begin{tabular}{p{0.45\textwidth} p{0.45\textwidth}}
    \textbf{Data} &  \textbf{Templates} \\ 
     \textit{Asterix (comicsCharacter) | creator | René Goscinny}  
     & The creator of Asterix (comicsCharacter) is René Goscinny. \\
     \textit{René Goscinny | nationality | French people} & René Goscinny was French people. \\
     
     \textit{Asterix (comicsCharacter) | creator | Albert Uderzo} & The creator of Asterix  (comicsCharacter) is Albert Uderzo. \\
     
    \rule{0pt}{3ex}\textbf{Text} \\  \multicolumn{2}{p{0.9\textwidth}}{asterix was created by rené goscinny and albert uderzo, the former  being a french national.}\\
    
     \rule{0pt}{3ex}\textbf{Human Output}  &  \textbf{NLI Output} \\  2.5 \textit{(=OK)} & \textit{omission} \\  
    \rule{0pt}{3ex}\textbf{Commentary} & \\
    
    \multicolumn{2}{p{0.9\textwidth}} {The sentence is OK, but the format of the values is not maintained in the text (``comicsCharacter'' is missing). The NLI treats this as an omission. Similar cases involve also e.g. number formatting.}
     \end{tabular}
     
    \label{tab:supp-webnlg-4}
\end{table}
    
\end{enumerate}

\newpage
\section*{E2E Error Analysis}

We checked 100 randomly sampled examples from the E2E data where our approach and the slot error script gave different results. We identified 34 cases where the slot error script was indeed correct, 45 where the script was incorrect but our NLI-based approach provided a correct result, and 18 cases where both annotations were incorrect or where it was not possible to unambiguously decide between the two.

The main error types identified, with counts and examples, are as follows (note that the analyzed examples may belong to multiple or none of the above classes):
\begin{enumerate}[label={\bf (\arabic*)},leftmargin=14pt,labelwidth=12pt,labelsep=2pt]

    \item \textbf{Value interpretation problems} -- 40 counts. Example:

\begin{table}[h]
    \centering \small
    \renewcommand{\arraystretch}{1.25}
    \begin{tabular}{p{0.45\textwidth} p{0.45\textwidth}}
    \textbf{Data} &  \textbf{Templates} \\ 
     \textit{The Punter | eat type | restaurant}  
     & The Punter is a restaurant. \\
     \textit{The Punter | food | Indian} & The Punter serves Indian. \\
     \textit{The Punter | price range | high} & The Punter is in the high price range. \\
      \textit{The Punter | rating | average}  
     & The Punter has average customer rating. \\
     \textit{The Punter | area | city centre} & The Punter is located in the city centre. \\
     \textit{The Punter | family friendly | no} & The Punter is not family-friendly. \\
     \textit{The Punter | near | Express by Holiday Inn} & The Punter is located near Express by Holiday Inn. \\
     
    \rule{0pt}{3ex}\textbf{Text} \\  \multicolumn{2}{p{0.9\textwidth}}{The Punter is a high priced, average rated, adult only Indian restaurant located near Express by Holiday Inn in the city centre.}\\
    
     \rule{0pt}{3ex}\textbf{Slot Error Script}  &  \textbf{NLI Output} \\  \textit{OK} & \textit{hallucination} \\  
    \rule{0pt}{3ex}\textbf{Commentary} & \\
    
    \multicolumn{2}{p{0.9\textwidth}} {The text uses ``adult only'' to verbalize \emph{family-friendly=no}, which is generally considered correct in the E2E dataset. However, the NLI treats this as hallucination (``adult only'' does not necessarily follow from ``is not family friendly'').}
     \end{tabular}
     
    \label{tab:supp-e2e-1}
\end{table}

    \item \textbf{Incorrect patterns in the slot error script} -- 33 counts. Example:

\begin{table}[h]
    \centering \small
    \renewcommand{\arraystretch}{1.25}
    \begin{tabular}{p{0.45\textwidth} p{0.45\textwidth}}
    \textbf{Data} &  \textbf{Templates} \\ 
     \textit{The Cricketers | eat type | restaurant}  
     & The Cricketers is a restaurant. \\
     \textit{The Cricketers | food | Chinese} & The Cricketers serves Chinese. \\
     \textit{The Cricketers | price range | cheap} & The Cricketers is in the cheap price range. \\
      \textit{The Cricketers | rating | average}  
     & The Cricketers has average customer rating. \\
     \textit{The Cricketers | area | riverside} & The Cricketers is located in the  riverside. \\
     \textit{The Cricketers | family friendly | yes} & The Cricketers is family-friendly. \\
     \textit{The Cricketers | near | All Bar One} & The Cricketers  is located near All Bar One. \\
     
    \rule{0pt}{3ex}\textbf{Text} \\  \multicolumn{2}{p{0.9\textwidth}}{Cheap Chinese food for all the family can be found at The Cricketers  restaurant, near All Bar One, in the riverside area. Average  ratings.}\\
    
     \rule{0pt}{3ex}\textbf{Slot Error Script}  &  \textbf{NLI Output} \\  \textit{omission} & \textit{OK} \\  
    \rule{0pt}{3ex}\textbf{Commentary} & \\
    
    \multicolumn{2}{p{0.9\textwidth}} {The slot error script considers \emph{family-friendly=yes} as missing -- it probably does not include the pattern ``for all the family''. NLI has no problems handling this.}
     \end{tabular}
     
    \label{tab:supp-e2e-2}
\end{table}

\newpage
    \item \textbf{Edge cases, hard to interpret} -- 18 counts. Example:

\begin{table}[h]
    \centering \small
    \renewcommand{\arraystretch}{1.25}
    \begin{tabular}{p{0.45\textwidth} p{0.45\textwidth}}
    \textbf{Data} &  \textbf{Templates} \\ 
     \textit{The Mill | eat type | restaurant}  
     & The Mill is a restaurant. \\
     \textit{The Mill | food | English} & The Mill serves English. \\
     \textit{The Mill | price range | moderate} & The Mill  is in the moderate price range. \\
      \textit{The Mill | rating | 3 out of 5}  
     & The Mill has 3 out of 5  customer rating. \\
     \textit{The Mill | area | riverside} & The Mill is located in the riverside. \\
     \textit{The Mill | family friendly | yes} & The  Mill is family-friendly. \\
     \textit{The Mill | near | Café Rouge} & The Mill is located near Café Rouge. \\
     
    \rule{0pt}{3ex}\textbf{Text} \\  \multicolumn{2}{p{0.9\textwidth}}{The Mill is a moderate restaurant that serves English food. Yes it  is kids-friendly. Its customer rating is 3 out of 5. It is located  in the riverside area near Café Rouge.}\\
    
     \rule{0pt}{3ex}\textbf{Slot Error Script}  &  \textbf{NLI Output} \\  \textit{omission} & \textit{OK} \\  
    \rule{0pt}{3ex}\textbf{Commentary} & \\
    
    \multicolumn{2}{p{0.9\textwidth}} {It is hard to interpret ``moderate restaurant'' as a correct verbalization of \emph{price\_range=moderate}. The NLI makes this assumption while the slot error script does not.}
     \end{tabular}
     
    \label{tab:supp-e2e-3}
\end{table}

    \item \textbf{Off-topic hallucinations} -- 8 counts. Example:

\begin{table}[h]
    \centering \small
    \renewcommand{\arraystretch}{1.25}
    \begin{tabular}{p{0.45\textwidth} p{0.45\textwidth}}
    \textbf{Data} &  \textbf{Templates} \\ 
     \textit{Giraffe | eat type | restaurant}  
     & Giraffe is a restaurant. \\
     \textit{Giraffe | food | English} & Giraffe serves English. \\
     \textit{Giraffe | area | riverside} & Giraffe is  located in the riverside. \\
      \textit{Giraffe | family friendly | yes}  
     & Giraffe is family-friendly. \\
     \textit{Giraffe | near | Rainbow Vegetarian Café} & Giraffe  is located near Rainbow Vegetarian Café. \\
     
    \rule{0pt}{3ex}\textbf{Text} \\  \multicolumn{2}{p{0.9\textwidth}}{Giraffe is a beautiful restaurant close to the Rainbow Vegetarian  Café. It is reasonably liked place serves English food and is  children friendly.}\\
    
     \rule{0pt}{3ex}\textbf{Slot Error Script}  &  \textbf{NLI Output} \\  \textit{omission} & \textit{hallucination+omission} \\  
    \rule{0pt}{3ex}\textbf{Commentary} & \\
    
    \multicolumn{2}{p{0.9\textwidth}} {While both the slot error script and the NLI detect the missing verbalization of \emph{area}, NLI probably interprets ``beautiful'' and ``reasonably liked'' as hallucinations, while the slot error script is not able to detect such cases.}
     \end{tabular}
     
    \label{tab:supp-e2e-4}
\end{table}

\end{enumerate}

\end{document}